\documentclass[letterpaper]{article}
\usepackage{aaai}
\usepackage{times}
\usepackage{helvet}
\usepackage{courier}
\usepackage{natbib}
\usepackage{graphicx}
\usepackage{float}
\usepackage{subcaption}
\floatstyle{ruled}
\usepackage[utf8]{inputenc}
\usepackage{color,soul}
\usepackage{algorithm}
\usepackage{algpseudocode}
\usepackage{multirow}
\usepackage[labelfont=bf]{caption}
\newfloat{pseudocode}{ht}{ext}
\floatname{pseudocode}{Algorithm}
\begin{document}
%
\graphicspath{ {./images/} }
\title{Exploring TD error as a heuristic for $\sigma$ selection in Q($\sigma, \lambda$)}
\author{Abhishek Nan\\
University of Alberta\\
Department of Computing Science\\
anan1@ualberta.ca\\
}
\maketitle
\begin{abstract}
\begin{quote}
In the landscape of TD algorithms, the Q($\sigma, \lambda$) algorithm is an algorithm with the ability to perform a multi-step backup in an online manner while also successfully unifying the concepts of sampling with using the expectation across all actions for a state. $\sigma \in [0,1]$ indicates the extent to which sampling is used. Selecting the value of $\sigma$ can be based on characteristics of the current state rather than having a constant value or being time based. This project explores the viability of such a TD-error based scheme.
\end{quote}
\end{abstract}

\section{Introduction}
 While having different dimensions of generalizability in an algorithm can serve as a powerful tool, in most cases it comes with the associated burden of having to manually select values along these dimensions, commonly referred to as hyper-parameter selection. In case of learning algorithms, an ideal algorithm would be completely general, even to the point that they do not need a fixed set of hyper-parameters for which they perform optimally for a given problem. In the context of Q($\sigma, \lambda$), the introduction of the $\sigma$ parameter gives us flexibility in terms of adjusting the proportion of sampling and expectation we want in our updates.
 But at the same time, while $\sigma$ does serve as a hyper-parameter, atypically a constant value of $\sigma$ was found to not have the best performance by  De Asis, Hernandez-Garcia, Holland and Sutton \citeyearpar{de2018multi}. They used a Dynamic Decay $\sigma$ scheme for n-step Q($\sigma$) where they reduced the value of $\sigma$ after every episode by a factor of 0.95. The intuition behind this being that initially using more of the expectation component would lead to poor results and hence, start off with a large sampling component and slowly decay towards using more of the expectation. This worked well in the environments they tried it on.

But potentially, there might be better schemes for $\sigma$ selection on a more granular level than the episode level and dynamically adjusting it based on factors of the environment/agent which could lead to better performance, especially in highly stochastic environments. For instance, Dynamic decay $\sigma$ is essentially an exponential decay, so depending on the stochasticity of the environment maybe a decay using an inverse square law might work better? Maybe a scheme which alternates the value of $\sigma$ between 0 (complete expectation) and 1 (complete sampling) for every episode or even for every time-step would work well? An approach based on the relative TD-errors between successive episodes might also serve as an indicator for the degree of sampling required. Of these alternatives, the last one performed the best in initial experiments and has been pursued further in this report.

A primary motivator for choosing Q($\sigma, \lambda$) rather than n-step Q($\sigma$) (which is introduced in the Sutton and Barto textbook) for this experiment was the idea that using the trace vectors as a heuristic for choosing $\sigma$ on a per step basis could work well. The intuition does not seem unreasonable that the more often a state is visited, the more accurate it's estimated state/action value(s) will become and the expectation component for that state can be used progressively more. Rather than storing the absolute number of visits, the trace vector maintains exactly this information in an indirect way. The trace for a state rises once it is visited and decays away as it is not visited for some time. The issue with this approach was that though it sounds intuitive for the tabular case, it might not be extended as intuitively for cases with function approximation. But this remains a line of potential future work.

\section{Selecting $\sigma$ based on TD-error}
Due to the inherent nature of reinforcement learning algorithms, initial episodes will have high TD error values and as time elapses, in subsequent episodes the value estimates move closer and closer to their true values; hence, the TD error value comes down. This could be used as a simple heuristic for deciding how good or bad the algorithm is performing in the current episode in terms of accuracy to the true values and then adjust the value of $\sigma$ accordingly for the following time-steps.

The basic idea is that initially $\sigma$ starts with a value of 1 for the first episode. For each subsequent episode, at every time step, $\sigma$ is calculated as the ratio of the current TD error and the maximum TD error from the previous episode. Of course, an alternative version of the algorithm could be framed using the average TD error instead of the maximum TD error as well; but in initial experiments, the maximum version performed better and is explored here. A potential issue with this intuition is that calculating the TD Error itself uses the current $\sigma$ value; so, after calculating the $\sigma$ value based on the TD Error, the new $\sigma$ value will come into effect only during the next time-step. But given that time-steps are obviously temporally sequential, it might not be too much of a stretch to suggest that visit frequencies of states in successive time-steps will be highly correlated and by extension so will their TD errors be as well. Hence, we can use the proposed intuition. Algorithm \ref{algorithm1} presents a version of Q($\sigma, \lambda$) with accumulating traces as presented by \cite{DBLP:journals/corr/abs-1711-01569}. The additional steps introduced in this project for a TD-error based $\sigma$ selection have been highlighted. The term \textit{TD Error Based $\sigma$} is used in this report to refer to the $\sigma$ selection scheme introduced in Algorithm \ref{algorithm1} and the term \textit{Dynamic Decay $\sigma$} is used to refer to the scheme used by \cite{de2018multi}.

\begin{algorithm}[!htbp]
\fontfamily{cmss}\selectfont
\begin{algorithmic}[1]
\State Initialize $Q(s, a) \quad \forall s \in \mathcal{S}, a \in \mathcal{A}$
\State \hl{$episode\_count \leftarrow 0$}
\State \hl{$\sigma \leftarrow 1$}
\For {each episode}

    \State $E(s, a) \leftarrow 0 \quad \forall s \in \mathcal{S}, a \in \mathcal{A}$
    \State Initialize $S_0 \neq$ terminal
    \State Choose $A_0$, e.g. $\epsilon$-greedy from $Q(S_0, .)$
    \State \hl{Initialize or Clear $TD\_error\_list$}
    \For {each step of episode}

        \State Take action $A_t$, observe $R_{t+1}$ and $S_{t+1}$
        \State Select $A_{t+1}$, e.g. $\epsilon$-greedy from $Q(S_{t+1}, .)$
        \State $V_{t+1}=\sum_{a'} \,\pi(a'|S_{t+1}) \; Q(S_{t+1}, a'))$
        \State $\delta = R_{t+1} + \gamma \, (\sigma \, Q(S_{t+1}, A_{t+1}) \newline
        \hspace*{5em}+ (1- \sigma) \; V_{t+1} - Q(S_{t}, A_t) $
        \State \hl{Add $\delta$ to $TD\_error\_list$}
        \State $E(S_t, A_t) \leftarrow E(S_t, A_t) + 1$
        \State $Q(s, a) \leftarrow Q(s, a) + \alpha \, \delta \, E(s, a)$
        \State $E(s, a) \leftarrow \gamma \lambda E(s, a) ( \sigma + \newline 
        \hspace*{5em}(1 - \sigma) \pi(A_{t+1} | S_{t+1}))$
        \State $A_t \leftarrow A_{t+1}$, $S_t \leftarrow S_{t+1}$
        \If {$episode\_count > 0$}
            \State \hl{$\sigma \leftarrow max(0,min(\mid\delta/\delta_{max}\mid,1))$}
        \EndIf
        \If {$S_t$ is terminal}
                \State \hl{$episode\_count \leftarrow episode\_count + 1$}
                \State \hl{$\delta_{max} \leftarrow max(TD\_error\_list$)}
                \State Break
        \EndIf
    \EndFor
\EndFor
\end{algorithmic}
\caption{Q($\sigma, \lambda$) with TD-error based $\sigma$ selection}
\label{algorithm1}
\end{algorithm}

\section{Empirical Evaluation}
\bigbreak
The environments selected for experiments all vary based on certain different characteristics which might be useful to evaluate on. These include the on-policy prediction and control settings, tabular representation and function approximation, stochastic and non-stochastic environments. In all experiments, the evaluation is a comparison between the performances of Dynamic Decay $\sigma$ and TD Error Based $\sigma$.

\section{Experiments and Results}
\subsection{19-State Random Walk}

\subsubsection{Environment}
\begin{figure}[!htbp]
\caption{The 19-State Random Walk MDP. The goal is to learn the true value of each state under equiprobable random behaviour. Source: \cite{de2018multi}}
\includegraphics[scale=0.5]{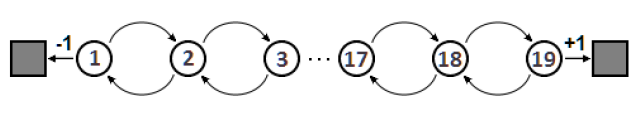}
\label{fig:19random}
\end{figure}

This is the same variation of the original 19-state random walk environment as used by \cite{de2018multi} (Figure \ref{fig:19random}). There are two terminal states at the ends; finishing at the state on the right gives a reward of +1 and the one on the left gives a reward of -1. All other transitions give a reward of 0. Unlike the original formulation in \cite{Sutton:1998:IRL:551283}, the transitions in this modified version are deterministic. So, each agent selects among 2 actions (left and right) in each non-terminal state and deterministically transitions to the next state in that direction.

\subsubsection{Agent Setup}
To evaluate the performance of both $\sigma$ variation schemes on prediction tasks, this modified version of the 19-State Random Walk was used. The agent in both cases learnt on-policy using an equiprobable random behaviour policy. There was no discounting ($\gamma=1$).

\subsubsection{Experiment}
For a comparative analysis using this environment, we looked at the plot of the RMS error (between the estimated value function and the analytically computed true values) over episodes elapsed. An initial hyper-parameter search was done for both algorithms, optimizing for the area under the said curve. $\lambda$ values of 0.1, 0.2, 0.3, 0.4, 0.5, 0.6, 0.7, 0.8, 0.9 with step-sizes ($\alpha$) of 0.001, 0.05 0.1, 0.2, 0.3, 0.4, 0.5, 0.6, 0.7, 0.8, 0.9 were evaluated over 10 runs. The best parameters for Dynamic decay $\sigma$ were found to be $\lambda=0.7$ and $\alpha=0.9$; and for TD error based $\sigma$, $\lambda=0.7$ and $\alpha=0.8$.

\subsubsection{Result}

\begin{figure}[!htbp]
\caption{RMS Error in the value function over 50 episodes of 19-State Random Walk. The results are averaged over 1000 runs. The lighter shades represent a 99\% confidence interval. Dynamic decay $\sigma$ had the better asymptotic solution and was the faster learner.}
\hspace*{-6mm}  
\includegraphics[scale=0.6]{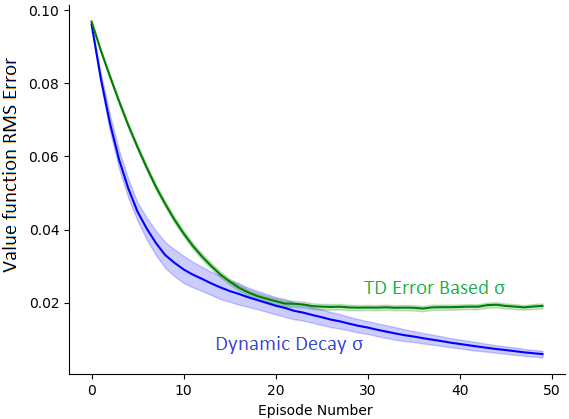}
\label{fig:19state_1}
\end{figure}

Figure \ref{fig:19state_1} shows the results of the best performing hyper-parameters for both schemes for 50 episodes averaged out over 1000 runs.  As can be seen, not only did the dynamic decay scheme perform better in terms of learning speed, it asymptotically approaches the true values of each state since the RMS error tends to 0. On the other hand, the RMS error for the TD error based scheme seems to plateau out after a point, which means it does not approach the true values asymptotically. 

\subsubsection{Analysis}
\begin{figure}[!htbp]
\caption{RMS Error in the value function over 50 episodes of 19-State Random Walk averaged over 10000 runs (lighter shades represent a 99\% confidence bound); TD Error Based + Dynamic Decay $\sigma$ refers to the TD error scheme while decaying by an additional factor of 0.95 per episode. This combined scheme has better asymptotic performance than plain Dynamic Decay $\sigma$, which proves that some sort of decaying scheme for $\sigma$ might be necessary in all environments, but at the same time a decay factor of 0.95 per episode is not always optimal.}
\hspace*{-6mm}  
\includegraphics[scale=0.6]{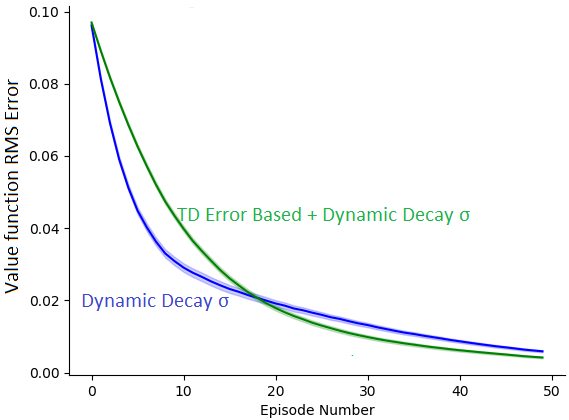}
\label{fig:19state_2}
\end{figure}

The results were a bit surprising, but still consistent with the previous observations by \cite{de2018multi}. Even in their experiments, only Dynamic decay $\sigma$ and Q(0) (i.e. Tree-backup) approach the true values asymptotically. This probably happens because we use a constant step-size and the estimates won't converge to a single value unless we reduce the step-size over time. In case of Q(0), it doesn't use sample updates, but rather updates in expectation and hence, does not suffer from this issue. Similarly, Dynamic decay $\sigma$ decays $\sigma$ over time to use more and more of the expectation component and at the same time gains a boost in initial performance by using sample updates to have a better overall performance than Q(0). But the key issue here is that there is no direct connection between the decay rate and the amount of sampling required. For instance, if the environment were much simpler, we would very likely have better performance by decaying $\sigma$ more sharply by a factor larger than 0.95. Even in this problem, we can achieve better asymptotic performance (Figure \ref{fig:19state_2}) by combining both TD error based and dynamic decay schemes; i.e. we use the normal TD error scheme as is, except we also multiply it by a factor of $0.95^{E_n-1}$, where $E_n$ is the episode number. The latter part is actually the episodic exponential decay as is used by Dynamic Decay $\sigma$. Conversely, there will be problems complicated enough in which the rate at which we build accurate estimates to bootstrap off does not correspond neatly to a decay of $\sigma$ by 0.95 per episode, and the agent might need to use more of the sampling component for a longer time and hence, a slower decay rate might be required. Furthermore, if the environment is stochastic, dynamic decay $\sigma$ might even lead to poor performance over time since it would be bootstrapping a large amount off estimates (as part of the expectation component) which would become inaccurate due to the stochasticity. A steadily decaying $\sigma$ will not be able to account for this stochasticity.

\subsection{Stochastic Windy Gridworld}

\subsubsection{Environment}
\begin{figure}[!htbp]
\caption{The Windy Gridworld environment}
\includegraphics[scale=0.73]{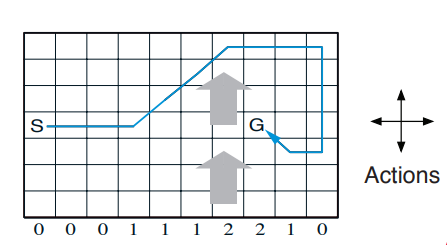}
\label{fig:windygrid}
\end{figure}

The Stochastic Windy Gridworld (SWG) is a variation of the original windy gridworld environment as described in \cite{Sutton:1998:IRL:551283} (Figure  \ref{fig:windygrid}). In this variation, the result of choosing an action is not deterministic. On choosing an action, the rules of state transition from the original environment carry over, except for the new effect that, 10\% of the time, the next state is a random selection from among the 8 states currently surrounding the agent. This environment was chosen to evaluate the performance of both $\sigma$ variation schemes under stochastic conditions.

\subsubsection{Agent Setup}
For all agents interacting with this environment, the discount factor had a fixed value of $\gamma$ = 1 and the behaviour policy was $\epsilon$-greedy with $\epsilon$ set to 0.1. Learning was done on-policy.
For this environment, Dynamic Decay $\sigma$ (decay by a factor of 0.99 at the end of every episode) is used as the benchmark and the new TD error based scheme is compared against it. 

\subsubsection{SWG Experiment 1}

\begin{figure}[!htbp]
\caption{Stochastic Windy GridWorld results for different decay rates in Dynamic Decay $\sigma$ with $\lambda=0.7$. The numbers in parenthesis represent the starting value and the decay factor of $\sigma$ after every episode. The plot shows the performance in terms of the average return per episode over 100 episodes as a function of the step-size. The results are averaged over 1000 runs for selected values, then are connected by straight lines. A decay rate of 0.99 had the best average return per episode across all step sizes.}
\hspace*{-6mm}  
\includegraphics[scale=0.58]{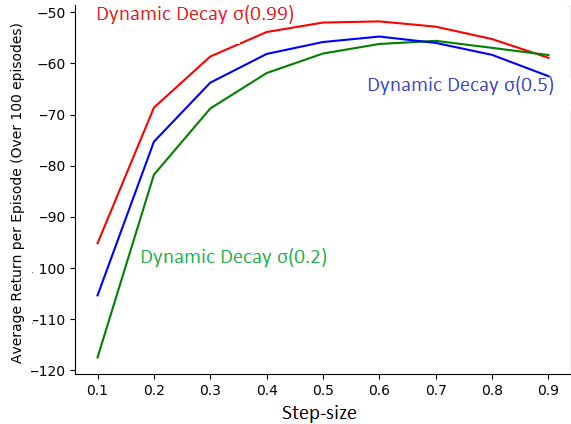}
\label{fig:misc1}
\end{figure}

Initially, just for sanity testing, a few simulations were ran with different values of the decaying factor (0.99, 0.95, 0.8, 0.5, 0.2) for dynamic decay $\sigma$ (Figure \ref{fig:misc1}) with $\lambda=0.7$. (This value of $\lambda$ was selected after a few initial experiments and its validity will be evident in the follow-up experiments). Figure \ref{fig:misc1} shows only a few representative plots to avoid cluttering. The best performance in terms of average return per episode over 100 episodes was observed for a decay factor of 0.99, which conforms to the observations of \cite{DBLP:journals/corr/abs-1711-01569} for Q($\sigma, \lambda$) on the Stochastic Windy Gridworld problem. Although, it should be noted that \cite{de2018multi} evaluated n-step Q($\sigma$) on the Stochastic Windy Gridworld problem and decay by a factor of 0.95 was found to be the best. This difference in the factor value can possibly be attributed to the different algorithms. But still, for the rest of the environments considered in this report, the benchmark is set to decaying $\sigma$ by a rate of 0.95 as that was the best performing benchmark found by \cite{de2018multi} and no such benchmark is available for Q($\sigma, \lambda$) in these environments.

\subsubsection{SWG Experiment 2}
For all further experiments for Stochastic Windy Gridworld, the decay factor in Dynamic Decay $\sigma$ was kept constant since a decay factor of 0.99 was found to be the best performer. The average return over 100 episodes was compared across different step-sizes, and also for different values of $\lambda$. $\lambda$ values of 0.1, 0.2, 0.3, 0.4, 0.5, 0.6, 0.7, 0.8 and 0.9 were tried out, but for a clear presentation, only a few representative $\lambda$ values are shown in the plots below.

\begin{figure}[!htbp]
\caption{Stochastic Windy GridWorld results for average return per episode over 100 episodes (averaged over 1000 trials) at different step-sizes and $\lambda$ values. Both $\sigma$ selection schemes had almost identical performances with the TD Error based scheme performing a little better at lower $\lambda$ values and Dynamic Decay $\sigma$ having slightly better  performance for $\lambda=0.7$.}
\hspace*{-7mm}  
\includegraphics[scale=0.58]{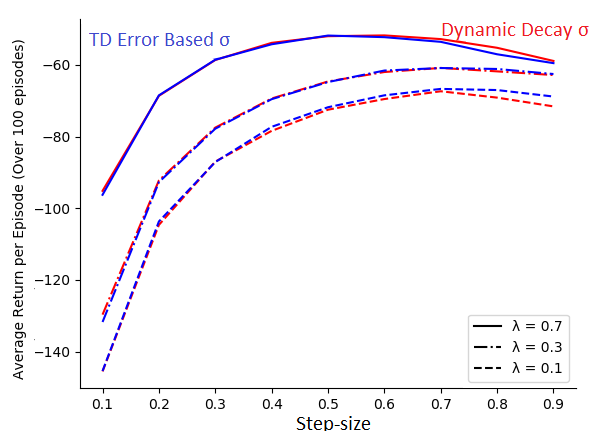}
\label{fig:graph4_1_max}
\end{figure}

\subsubsection{Results}
Trials run with the TD Error Based $\sigma$ (Figure \ref{fig:graph4_1_max}) performed comparably with Dynamic Decay $\sigma$. It outperformed  Dynamic Decay $\sigma$ at lower $\lambda$ values for most step-sizes and came close to the peak performance of Dynamic Decay $\sigma$ at $\lambda=0.7$, albeit losing out by a very small margin (0.18). The standard errors were all less than 0.5.

\subsubsection{Analysis}
Since peak performance is an important concern, this warrants a further investigation of both $\sigma$ selection schemes. But the results so far do serve as a positive indicator that a TD error based $\sigma$ value can indeed be viable.

\subsubsection{SWG Experiment 3}
\begin{figure}[!htbp]
\caption{Stochastic Windy Gridworld performances for both $\sigma$ variation schemes at $\lambda=0.7$ and $\alpha=0.5$. The plot shows the return per episode (averaged over 1000 trials) past the first 5 episodes. The lighter shades represent a 99\% confidence interval. Dynamic Decay $\sigma$ performed better in terms of the final asymptotic solution.}
\hspace*{-6mm}  
\includegraphics[scale=0.6]{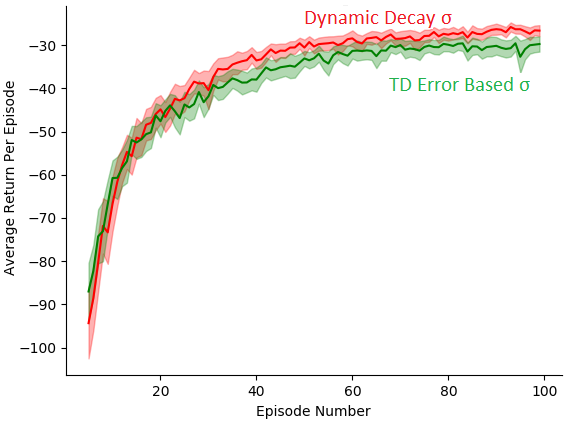}
\label{fig:graph4_2}
\end{figure}

Figure \ref{fig:graph4_2} presents a closer look at the performance of both algorithms at their best hyper-parameters as found in Figure \ref{fig:graph4_1_max}.

\subsubsection{Results}
Dynamic decay $\sigma$ did indeed perform better in terms of the asymptotic solution by a small, but statistically significant margin as we saw in the previous experiment also. The speed of learning in both cases is very similar.

\subsubsection{Analysis}
The result was somewhat surprising since using a customised $\sigma$ selection scheme, the expectation was that it would lead to faster learning performance, but to the same asymptotic solution, if not better. A possible explanation for this is that the TD Error Based scheme does not decay the sampling component to 0 and hence, this leads to high variance updates even at later time-steps where the Dynamic Decay $\sigma$ scheme has annealed $\sigma$ and majorly uses the expectation component for better performance. This leads us on to our next experiment where such a behaviour might not be optimal.

\subsection{Moving Goal Windy Gridworld}

\subsubsection{Environment}

This environment is exactly like the original windy gridworld from \cite{Sutton:1998:IRL:551283}, except the end goal state changes every 10 episodes. There is no stochasticity involved in the transitions as in the stochastic windy gridworld environment described before.

The motivation behind the formulation of this variation of the Windy Gridworld problem is the fact that in dynamic decay $\sigma$, the $\sigma$ value constantly decays with no way of recovering a higher sampling component later if the environment is stochastic. During long sequences of episodes, the $\sigma$ value becomes very small and the expectation component takes over; but if the environment has changed, then this will lead to poorer performance as compared to a $\sigma$ variation scheme which can adjust it's value depending on the accuracy of the current estimates using the TD error as a relative measure.

\subsubsection{Agent setup}
The setup is exactly same as the Stochastic Windy Gridworld, i.e. $\gamma=1$ and the behaviour policy was $\epsilon$-greedy with $\epsilon$ set to 0.1. Learning was done on-policy.

\subsubsection{Experiment}
Each run comprises of 100 episodes and the goal state changes every 10 episodes. Initially a hyper-parameter search was done with $\lambda$ values of 0.1, 0.2, 0.3, 0.4, 0.5, 0.6, 0.7, 0.8, 0.9 and step-sizes ($\alpha$) of 0.001, 0.01, 0.1, 0.2, 0.3, 0.4, 0.5, 0.6, 0.7, 0.8, 0.9. Optimizing for the total return over 100 episodes, the best results for Dynamic Decay $\sigma$ was found to be $\lambda=0.8, \alpha=0.6$ and for TD-error based $\sigma$, $\lambda=0.6, \alpha=0.8$.

\begin{figure}[!htbp]
\caption{Moving Goal Windy Gridworld results (averaged over 1000 runs) for 100 episodes with the end goal state changing every 10 episodes. The TD Error based $\sigma$ selection scheme performed better in terms of the total return over 100 episodes.}
\hspace*{-6mm}  
\includegraphics[scale=0.6]{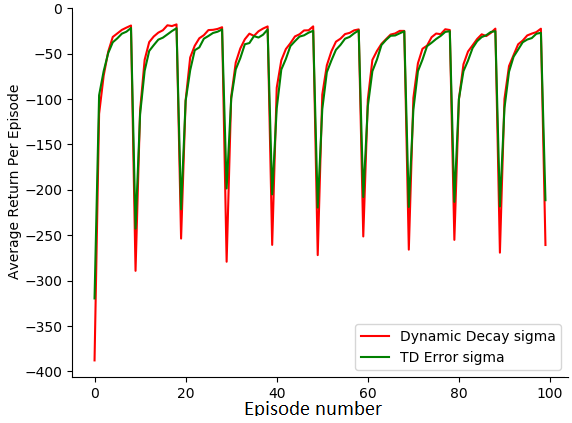}
\label{fig:moving_goal1}
\end{figure}

\subsubsection{Results}

Figure \ref{fig:moving_goal1} presents the results of this experiment. The total return over the entire 100 episodes (averaged over 1000 trials) was calculated. For Dynamic Decay $\sigma$ the total return was -6766.92 (Standard Error = 15.26) and for TD error based $\sigma$ the total was -6514.43 (Standard Error = 38.78). This meant that the TD error scheme had a better overall performance in terms of the total return. The standard errors imply that the results were statistically significant with a confidence $>$ 99\%.

\subsubsection{Analysis}
As is visible in Figure \ref{fig:moving_goal1} there seems to be a pattern here that whenever the goal state changes the performance of the TD error based $\sigma$ scheme does not deteriorate as much as it does in case of Dynamic Decay $\sigma$; albeit they converge to approximately the same solution before the goal changes again. The intuition behind TD error based $\sigma$ seems to be a valid one. Over time, if the environment is stochastic enough, this scheme can perform better than Dynamic decay $\sigma$. The effect demonstrated in this experiment should be more evident for longer trials due to the expectation component becoming larger and larger in Dynamic Decay $\sigma$ as time goes on. Similarly, in environments with larger action spaces (since the expectation component depends on all actions) this effect should be exacerbated.

\subsection{MountainCar}

\subsubsection{Environment}

\begin{figure}[!htbp]
\caption{The MountainCar environment}
\includegraphics[scale=0.53]{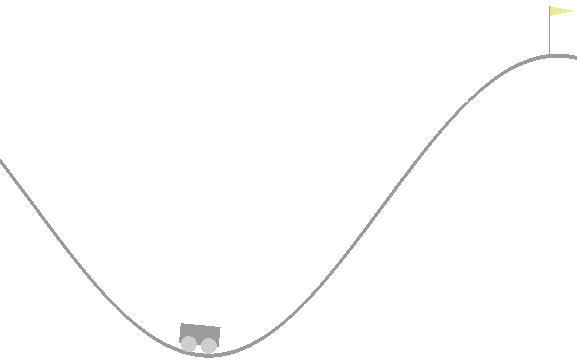}
\label{fig:mountaincar}
\end{figure}

The Mountain Car environment in the experiments performed here is used exactly as is formulated in \cite{Sutton:1998:IRL:551283}. 

\subsubsection{Agent Setup}
This environment is considered in order to evaluate the two $\sigma$ schemes in question on problems which require function approximation. No discounting was done, i.e. $\gamma=1$ and the behaviour policy was $\epsilon$-greedy with $\epsilon$ set to 0.1. Learning was done on-policy.

Since the state space is continuous in this problem, function approximation was done via version 3 of Sutton's tile coding implementation. 8 tilings were used with each tiling being offset from the previous by consecutive odd integers.

\subsubsection{Experiment}

For the MountainCar environment, we look at the average return per episode. Each episode runs for a maximum of 3000 steps and then terminates (to prevent cases of divergence going until infinity). Here the benchmark is Dynamic Decay $\sigma$ by a factor of 0.95 as reported by \cite{de2018multi}.

A hyper-parameter grid search was done for Dynamic Decay $\sigma$ with 100 runs each. For each simulation, step-size ($\alpha$) values of 0.9, 0.5, 0.25, 0.1, 0.05, 0.001 were considered and $\lambda$ values of 0.1, 0.2, 0.5, 0.7, 0.9. Optimizing for the return after 500 episodes, the best values for the hyper-parameters were found to be $\lambda=0.1$ and $\alpha=0.5$. The same hyper-parameters were the best performers as well when using the TD error based sigma scheme.
\subsubsection{Results}
\begin{figure}[!htbp]
\caption{MountainCar results showing the average return per episode (averaged over 500 trails) starting from 50 episodes uptil 300 episodes. The lighter shades represent a 95\% confidence interval. TD Error Based $\sigma$ performed comparably in terms of the initial learning speed and approaches a better asymptotic solution over time.}
\hspace*{-6mm}  
\includegraphics[scale=0.6]{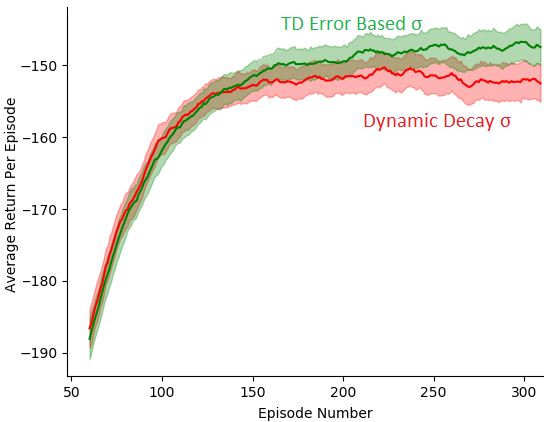}
\label{fig:mountaincar_1}
\end{figure}
Figure \ref{fig:mountaincar_1} shows the averaged results of 500 runs of MoutainCar with Dynamic Decay $\sigma$ and TD error based $\sigma$ selection. For initial episodes, the performance of both schemes was very close and so, in order to provide a better comparison the graph presents the results after 50 episodes. As, can be seen, the TD error based scheme was the better performer in terms of the asymptotic solution.
\subsubsection{Analysis}
Since this is a problem with continuous state spaces, a customised per step $\sigma$ value can potentially lead to better performance as we see in the results. Although the intuition was that we would see better performance in terms of the learning speed, we actually ended up seeing very similar learning speeds, but better asymptotic performance from the TD based scheme. From this it could be inferred that results of a per state $\sigma$ selection scheme would be more evident in complex problems (as compared to distinct state gridworld problems, where a simple constant $\sigma$ decay might work well enough) which require function approximation. So, we try the same experiments on another continuous state space problem which requires function approximation.

\subsection{CartPole}
\subsubsection{Environment}

\begin{figure}[!htbp]
\caption{The CartPole environment}
\includegraphics[scale=0.53]{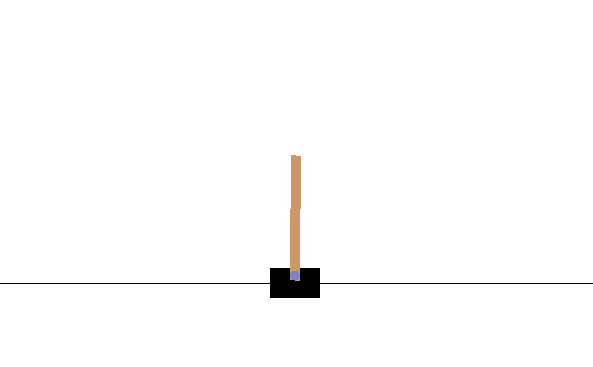}
\label{fig:cartpole_exp}
\end{figure}

The CartPole environment in the experiments performed here is used exactly as is formulated in \cite{Sutton:1998:IRL:551283}.

\subsubsection{Agent Setup}
Here we compare two agents using the two different $\sigma$ selection schemes. There is no discounting, i.e. $\gamma=1$ and the behaviour policy was $\epsilon$-greedy with $\epsilon$ set to 0.1. Learning was done on-policy.

Since the state space is continuous in this problem, function approximation was done via version 3 of Sutton's tile coding implementation. 8 tilings were used with each tiling being offset from the previous by consecutive odd integers.

\subsubsection{CartPole Experiment 1}
For CartPole, we look at the return per episode. Initial experiments consisted of 30 runs, with each lasting 200 episodes. For each simulation, step-size ($\alpha$) values of 0.9, 0.5, 0.25, 0.1, 0.05, 0.001 and $\lambda$ values of 0.1, 0.2, 0.5, 0.7, 0.9 were considered. The best hyper-parameters based on the area under the curve (AUC) of the return per episode vs episode number plot (similar to Figure \ref{fig:cartpole_1}) for dynamic decay with a factor of 0.95 was evaluated to be $\alpha=0.5$ and $\lambda=0.7$. Similarly, the best performing hyper-parameters for the TD error based $\sigma$ scheme was also found to be $\alpha=0.5$ and $\lambda=0.7$.

For the actual evaluation, both Dynamic Decay $\sigma$ and TD error based $\sigma$ schemes were run with their respective best hyper-parameters.

\subsubsection{Results}

\begin{figure}[!htbp]
\caption{CartPole results for 200 episodes showing the return per episode (averaged over 100 trials). The results have been smoothed using a right-centered moving average with a window of 30 episodes. Lighter shades present a 70\% confidence interval. Although not statistically significant, the initial impression was that the TD Error based scheme had better performance.}
\hspace*{-6mm}  
\includegraphics[scale=0.6]{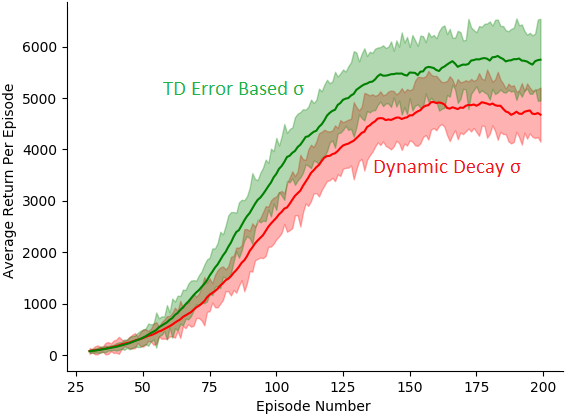}
\label{fig:cartpole_1}
\end{figure}

Figure \ref{fig:cartpole_1} presents the results of this experiment. The CartPole environment had a lot of variance and hence, the standard error was too high to be able to draw any significant conclusions. The number of trials was limited to 100, since each run of 200 episodes took quite some time.

\subsubsection{Analysis}
Despite the high variance, the trend seemed encouraging for TD Error Based $\sigma$, both in terms of asymptotic performance as well as in terms of learning speed. 

\subsubsection{CartPole Experiment 2}
So, as a follow up to the previous experiment more runs were done with fewer episodes but more iterations in an attempt to have more statistically significant results at least for the first few episodes. The number of runs was increased to 10000 and the number of episodes was restricted to 50. Every other hyper-parameter was the same as in the previous experiment.

\subsubsection{Results}
\begin{figure}[!htbp]
\caption{The results (averaged over 10000 trials) from repeating the same experiment in Figure \ref{fig:cartpole_1} uptil only 50 episodes. Lighter shades represent a 99\% confidence interval. TD Error Based $\sigma$ had the better performance  (faster learning) over the initial 50 episodes.}
\hspace*{-6mm}  
\includegraphics[scale=0.58]{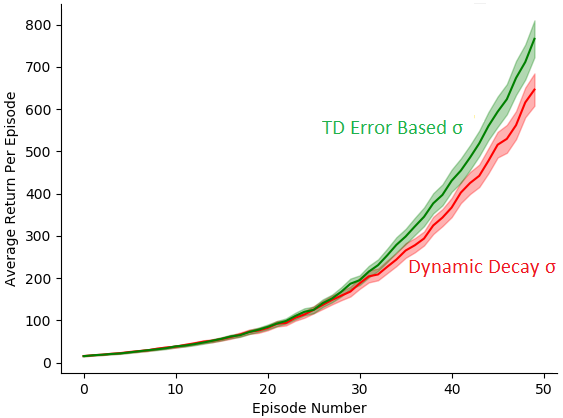}
\label{fig:cartpole_2}
\end{figure}

Figure \ref{fig:cartpole_2} presents the results for this experiment. The results haven't been smoothed for this particular experiment and are the actual average returns.
\subsubsection{Analysis}
From Figure \ref{fig:cartpole_2}, it would be fair to draw the conclusion that for the CartPole environment, using the TD error as a heuristic for selecting the value of $\sigma$ allows Q($\sigma,\lambda$) to perform better than dynamically decaying it by a constant factor of 0.95. It very likely also leads to better asymptotic performance, but the results from the experiment performed in Figure \ref{fig:cartpole_1} were not  very statistically significant.

\begin{figure}[!htbp]
\caption{The value of $\sigma$ as a function of the episode number in the CartPole environment for the two different $\sigma$ variation schemes in two independent runs. The decay pattern obtained in case of TD Error Based $\sigma$ led to better performance.}
\hspace*{-6mm}  
\includegraphics[scale=0.58]{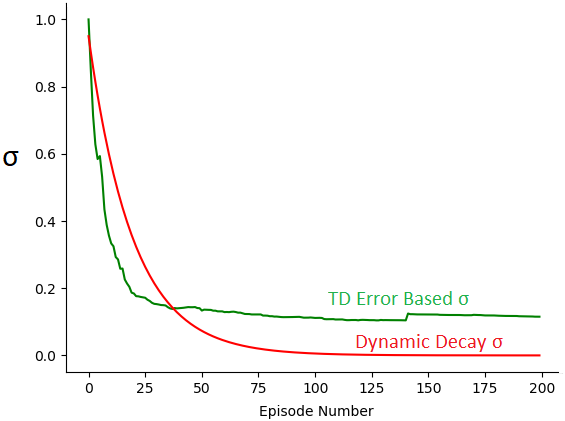}
\label{fig:cartpole_sigma}
\end{figure}

Furthermore, we look at the value of $\sigma$ per episode in both schemes over a sample run. From Figure \ref{fig:cartpole_sigma} we see how using the TD error naturally leads to a decay of the $\sigma$ value as well, but a slower decay overall than the scheme using a decay factor of 0.95 per episode.

This leads credence to our earlier hypothesis that the accuracy of the value estimates in every environment won't correspond neatly with a $\sigma$ decay rate by a constant value (0.95 in this case). A slower, faster or totally dynamic $\sigma$ variation scheme might be required for different environments.

\subsection{Conclusion}

Overall, there are a few conclusions we can draw from these experiments. It's quite difficult to claim either of the $\sigma$ selection schemes as the clear winner. In cases where the environment is simple enough (eg.windy gridworld), an episodic exponentially decaying scheme of $\sigma$, such as Dynamic Decay $\sigma$ will perform well due to the inherent nature of reinforcement learning algorithms; i.e. value estimates get better over time and being able to use the expectation over them will speed up learning. Also, due to its steadily decaying nature, it's likely to arrive at asymptotically better solutions in such environments (19-State Random Walk). But, in cases where the environment is stochastic, a constant decay of $\sigma$ will not correlate directly with the current level of the value estimates' correctness and hence, this might lead to sub-optimal performance. The level of estimate correctness is a difficult thing to quantify, and the relative TD error of successive episodes explored here was one such possible solution. It performs relatively well in a lot of scenarios, especially if function approximation is involved (eg. MountainCar, CartPole) or when environments are highly stochastic (eg. Moving goal windy gridworld).
A possible combination of both schemes might have suitable properties which account for the deficiencies of either $\sigma$ selection scheme individually. For instance, as we saw in Figure \ref{fig:19state_2} that ensuring some form of decay with the TD-error scheme might be necessary to get better asymptotic performance in certain environments (for eg. the 19-State Random Walk). This way we might be able to combine the (occasionally better) asymptotic performance of Dynamic Decay $\sigma$ with the benefits of TD Error Based $\sigma$.

\bibliography{citefile}
\end{document}